\documentclass{article}
\usepackage{spconf,amsmath,graphicx,multirow}

\title{Pseudo Labeling and Negative Feedback Learning for Large-scale Multi-label Domain Classification}
\name{Joo-Kyung Kim \qquad Young-Bum Kim}
\address{Amazon Alexa AI}
\begin{document}
%
\maketitle
\begin{abstract}
In large-scale domain classification, an utterance can be handled by multiple domains with overlapped capabilities.
However, only a limited number of ground-truth domains are provided for each training utterance in practice while knowing as many as correct target labels is helpful for improving the model performance. 
In this paper, given one ground-truth domain for each training utterance, we regard domains consistently predicted with the highest confidences as additional pseudo labels for the training.
In order to reduce prediction errors due to incorrect pseudo labels, we leverage utterances with negative system responses to decrease the confidences of the incorrectly predicted domains.
Evaluating on user utterances from an intelligent conversational system, we show that the proposed approach significantly improves the performance of domain classification with hypothesis reranking.
\end{abstract}
\section{Introduction}
Domain classification is a task that predicts the most relevant domain given an input utterance \cite{Tur2011}.\footnote{A domain is usually defined as an application or functionality than can handle specific intents \cite{Tur2011,Sarikaya2016,YBKim2018a}.}
It is becoming more challenging since recent conversational interaction systems such as Amazon Alexa, Google Assistant, and Microsoft Cortana support more than thousands of domains developed by external developers \cite{Kumar2017,YBKim2018a,YBKim2018b}. As they are independently and rapidly developed without a centralized ontology, multiple domains have overlapped capabilities that can process the same utterances. For example, ``\textit{make an elephant sound}'' can be processed by \texttt{AnimalSounds}, \texttt{AnimalNoises}, and \texttt{ZooKeeper} domains.

Since there are a large number of domains, which are even frequently added or removed, it is infeasible to obtain all the ground-truth domains of the training utterances, and domain classifiers for conversational interaction systems are usually trained given only a small number (usually one) of ground-truths in the training utterances. This setting corresponds to multi-label positive and unlabeled (PU) learning, where assigned labels are positive, unassigned labels are not necessarily negative, and one or more labels are assigned for an instance \cite{Kanehira2016,Yu2014}.\footnote{\cite{Kanehira2016} utilizes pairwise label dependencies whose computational complexity is a polynomial of degree 2 in terms of the number of labels, which is unsuitable in large-scale domain classification. \cite{Yu2014} is dealing with the scalability issue but they assume optimizing a low-rank linear model whose expressive power is limited.}

In this paper, we utilize user log data, which contain triples of an utterance, the predicted domain, and the response, for the model training. Therefore, we are given only one ground-truth for each training utterance. In order to improve the classification performance in this setting, if certain domains are repeatedly predicted with the highest confidences even though they are not the ground-truths of an utterance, we regard the domains as additional pseudo labels.
This is closely related to pseudo labeling \cite{Lee2013} or self-training \cite{Yarowsky1995,McClosky2006,Ruder2018}. While the conventional pseudo labeling is used to derive target labels for unlabeled data, our approach adds pseudo labels to singly labeled data so that the data can have multiple target labels.
Also, the approach is related to self-distillation, which leverages the confidence scores of the non-target outputs to improve the model performance \cite{Hinton2014,Furlanello2018}. While distillation methods utilize the confidence scores as the soft targets, pseudo labeling regards high confident outputs as the hard targets to further boost their confidences. We use both pseudo labeling and self-distillation in our work.
\begin{figure}[t]
	\centering
	\includegraphics[width=0.45\textwidth]{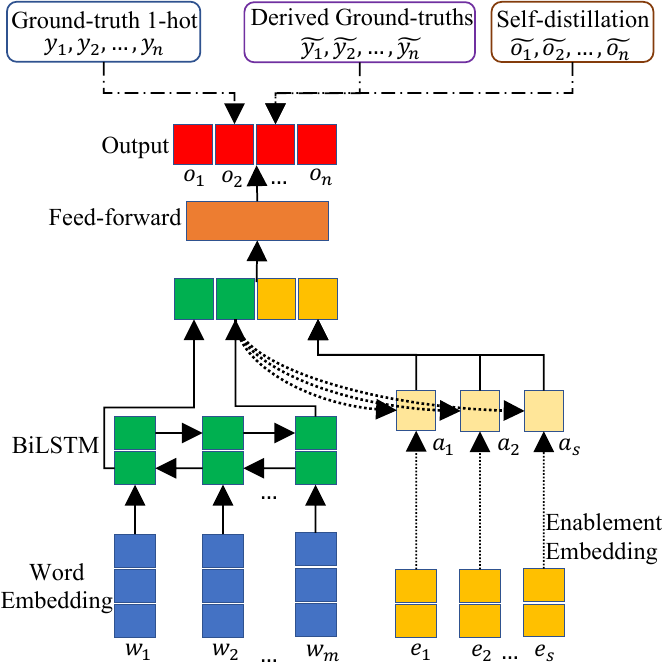}
\vspace{-2mm}
	\caption{\small Shortlister architecture: an input utterance is represented as a concatenation of the utterance vector from BiLSTM and the weighted sum of domain enablement vectors through domain enablement attention mechanism. Then, a feed-forward neural network followed by sigmoid activation represents the $n$-dimensional output vector.}
	\label{fig:shortlister_icassp20}
\vspace{-4mm}
\end{figure}

Pseudo labels can be wrongly derived when irrelevant domains are top predicted, which can lead the model training with wrong supervision.
To mitigate this issue, we leverage utterances with negative system responses to lower the prediction confidences of the failing domains.
For example, if a system response of a domain for an input utterance is ``\texttt{\textit{I don't know that one}}'', the domain is regarded as a negative ground-truth since it fails to handle the utterance.

Evaluating on an annotated dataset from the user logs of a large-scale conversation interaction system, we show that the proposed approach significantly improves the domain classification especially when hypothesis reranking is used \cite{Robichaud2014, YBKim2018b}.

\section{Model Overview}
We take a hypothesis reranking approach, which is widely used in large-scale domain classification for higher scalability \cite{Robichaud2014,YBKim2018b}.
Within the approach, a shortlister, which is a light-weighted domain classifier, suggests the most promising $k$ domains as the hypotheses. We train the shortlister along with the added pseudo labels, leveraging negative system responses, and self-distillation, which are described in Section \ref{sec:domain_classification}.
Then a hypothesis reranker selects the final prediction from the $k$ hypotheses enriched with additional input features, which is described in Section \ref{sec:reranking}.



\section{Shortlister Model}
\label{sec:domain_classification}
Our shortlister architecture is shown in Figure \ref{fig:shortlister_icassp20}.
The words of an input utterance are represented as contextualized word vectors by bidirectional long short-term memory (BiLSTM) on top of the word embedding layer \cite{Graves2005}. Then, the concatenation of the last outputs of the forward LSTM and the backward LSTM is used to represent the utterance as a vector.\footnote{In our experiments, using convolution neural networks \cite{Kim2014} or self-attention \cite{Lin2017} for encoding do not make significant differences.} Following \cite{YBKim2018a} and \cite{JKKim2018b}, we leverage the domain enablement information\footnote{Enabled domains are favorite or authenticated domains.} through attention mechanism \cite{Bahdanau2015}, where the weighted sum of enabled domain vectors followed by sigmoid activation is concatenated to the utterance vector for representing a personalized utterance. On top of the personalized utterance vector, a feed-forward neural network followed by sigmoid activation is used to obtain $n$-dimensional output vector $o$, where the prediction confidence of each domain is represented as a scalar value between 0 and 1.

Given an input utterance and its target label, binary cross entropy is used as the baseline loss function as follows:
\vspace{-3mm}
\begin{equation}
\mathcal{L}_{b} = -\sum_{i=1}^{n} y_i \log o_i + \left(1-y_i\right) \log \left(1-o_i\right),
\label{eq:base}
\vspace{-1mm}
\end{equation}
where $o$, $y$, and $n$ denote the model output vector, the one-hot vector of the target label, and the number of total labels.
We describe other proposed loss functions in the following subsections.

\subsection{Deriving Pseudo Labels}
\label{ssec:pseudo_labeling}
We hypothesize that the outputs repeatedly predicted with the highest confidences are indeed correct labels in many cases in multi-label PU learning setting.
This approach is closely related to pseudo labeling \cite{Lee2013} or self-training \cite{Yarowsky1995,McClosky2006,Ruder2018} in semi-supervised learning since our model is supervised with additional pseudo labels, but differs in that our approach assigns pseudo labels to singly labeled train sets rather than unlabeled data sets.

We derive the pseudo labels when the following conditions are met:
\begin{itemize}
    \item Maximally $p$ domains predicted with the highest confidences that are higher than the confidence of the known ground-truth.
    \item Domains predicted with the highest confidences for $r$ times consecutively so that consistent top predictions are used as pseudo labels.
\end{itemize}
For the experiments in Section \ref{sec:experiments}, we use $p$=2 and $r$=4, which show the best dev set performance.
Those derived pseudo labels are used in the model training as follows:
\vspace{-2mm}
\begin{equation}
\mathcal{L}_{d} = -\sum_{i=1}^{n} \tilde{y_i} \log o_i + \left(1-\tilde{y_i}\right) \log \left(1-o_i\right),
\label{eq:pseudo}
\end{equation}
where $\tilde{y}$ denotes an $n$-hot vector such that the elements corresponding to the original ground-truth and the additional pseudo labels are set to 1.

\subsection{Leveraging Negative Feedback}
\label{ssec:neg_feed_learn}
During the model training, irrelevant domains could be top predicted, and regarding them as additional target labels results in wrong confirmation bias \cite{Tarvainen2017}, which causes incorrect model training.
To reduce the side effect, we leverage utterances with negative responses in order to discourage the utterances' incorrect predictions.
This setting can be considered as a multi-label variant of Positive, Unlabeled, and Biased Negative Data (PUbN) learning \cite{Hsieh2019}.

We obtain training utterances from log data, where utterances with positive system responses are used as the positive train set in Equation \ref{eq:base} and \ref{eq:pseudo} while the utterances with negative responses are used as the negative train set in Equation \ref{eq:neg}.
For example, \texttt{AnimalSounds} is a (positive) ground-truth domain for ``\textit{a monkey sound}'' because the system response to the utterance is ``\textit{\texttt{Here comes a monkey sound}}'' while it is a negative ground-truth for ``\textit{a dragon sound}'' as the response is ``\textit{\texttt{I don't know what sound a dragon makes}}''.\footnote{Recent intelligent conversational systems can support thousands of domains, many of which are with very specific/narrow capabilities. For ``\textit{a dragon sound}'', \texttt{DungeonSound} and \texttt{DragonFire} are the correct domains, and predicting \texttt{ZooKeeper} is incorrect.}
\footnote{Negative responses can be easily identified since the responses are generated from predefined templates. We use 2K template patterns to extract such responses.
}

Previous work \cite{Weston2016, Hancock2019} excludes such negative utterances from the training set. We find that it is more effective to explicitly demote the prediction confidences of the domains resulted in negative responses if they are top ranked.
It is formulated as a loss function:
\begin{equation}
\mathcal{L}_{n} = 
\left\{\begin{matrix}
- \log \left(1-o_j\right) & \forall {i\neq j}\ o_i \leq  o_j\\ 
0 & \mathrm{Otherwise,}
\end{matrix}\right.
\label{eq:neg}
\end{equation}
where $j$ denotes the index corresponding to the negative ground-truth domain. We demote the confidences of the negative ground-truths only when they are the highest so that the influence of using the negative ground-truths is not overwhelming.\footnote{In our experiments, reducing confidences of negative ground-truths regardless of the confidence ranks shows worse performance.}

\subsection{Self-distillation}
Knowledge distillation has been shown to improve the model performance by leveraging the prediction confidence scores from another model or from previous epochs \cite{Hinton2014,Furlanello2018,JKKim2018b}. Inspired by \cite{JKKim2018b}, we utilize the model at the epoch showing the best dev set performance before the current epoch to obtain the prediction confidence scores as the soft target. The self-distillation in our work can be formulated as follows:
\vspace{-2mm}
\begin{equation}
\mathcal{L}_{s} = -\sum_{i=1}^{n} \tilde{o_i} \log o_i + \left(1-\tilde{o_i}\right) \log \left(1-o_i\right),
\vspace{-1mm}
\end{equation}
where $\tilde{o_i}$ denotes the model output at the epoch showing the best dev set performance so far. Before taking sigmoid to obtain $\tilde{o_i}$, we use 16 as the temperature to increase the influence of distillation \cite{Hinton2014}, which shows the best dev set performance following \cite{JKKim2018b}.

\subsection{Combined Loss}
The model is optimized with a combined loss function as follows:
\vspace{-2mm}
\begin{equation}
\mathcal{L} = \left(1-\alpha^t\right)\mathcal{L}_{b} + \alpha^t\left(\mathcal{L}_{d} + \mathcal{L}_{s}\right) + \beta\mathcal{L}_{n}.
\end{equation}
where $\alpha^t=1-0.95^t$ and $t$ is the current epoch so that the baseline loss is mainly used in the earlier epochs while the pseudo labels and self-distillation are more contributing in the later epochs following \cite{Hu2016}. $\beta$ is a hyperparameter for utilizing negative ground-truths, which is set to 0.00025 showing the best dev set performance.

\section{Hypothesis Reranking Model}
\label{sec:reranking}
\begin{figure}[t]
	\centering
	\includegraphics[width=0.3\textwidth]{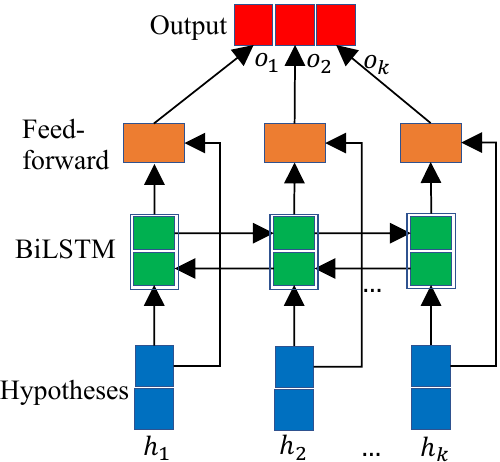}
\vspace{-2mm}
	\caption{\small Hypothesis Reranker Architecture: Each hypothesis consists of scores and vectors of domain, intent, and slots. Then, BiLSTM and a feed-forward neural network are used to represent contextualized hypothesis confidence scores.}
	\label{fig:hyprank_model}
\vspace{-4mm}
\end{figure}

\begin{table*}[ht]
\small
\centering
\begin{tabular}{l|l|llll|lll}
\multirow{2}{*}{} & \multirow{2}{*}{Model}                      & \multicolumn{4}{c|}{Shortlister}                            & \multicolumn{3}{c}{Hypothesis Reranker}         \\ \cline{3-9} 
                  &                                             & Precision      & Recall         & F-1            & nDCG$_3$           & Precision      & Recall         & F-1            \\ \hline
(1)                 & Base                                        & 77.27          & \textbf{83.27} & 80.15          & 71.92          & 79.13          & 82.34          & 80.71          \\
(2)                 & Base+pseudo                               & 76.77          & 82.87          & 79.70          & 71.64          & 79.02          & 81.23          & 80.11          \\
(3)                 & Base+neg\_feed                              & 77.90          & 81.32          & 79.58          & 72.15          & 79.33          & 83.54          & 81.38          \\
(4)                 & Base+neg\_feed+self\_dist                   & 77.73          & 82.46          & 80.03          & 72.24          & 79.24          & 83.71          & 81.41          \\
(5)                 & Base+pseudo+neg\_feed                     & \textbf{78.14} & 82.87          & 80.43          & 72.53          & \textbf{79.52} & 83.89          & 81.65          \\
(6)        & Base+pseudo+neg\_feed+self\_dist & 77.96      & 83.21          & \textbf{80.50} & \textbf{72.68} & 79.41          & \textbf{84.09} & \textbf{81.69} \\ \hline
\end{tabular}
\vspace{-2mm}
\caption{Evaluation results on various metrics (\%). pseudo, neg\_feed, and self\_dist denote using derived pseudo labels, negative feedback, and self-distillation, respectively.}
\label{tab:experiments}
\end{table*}

\begin{table*}[ht]
\small
\centering
\begin{tabular}{l|l|l}
Utterance                           & Known ground-truth  & Additional pseudo labels\\ \hline
One hundred twenty beats per minute       & Acoustic Metronome  & My Metronome, Metronome Lite\\ \hline
Play ocean sounds & Ambient Sounds & Sleep and Relaxation Sounds, Sleep Sounds: Ocean Sounds\\ \hline
Give me the news briefing & CBS News & The Washington Post, CNN\\ \hline
\end{tabular}
\vspace{-2mm}
\caption{Examples of additional pseudo labels.}
\label{tab:examples}
\end{table*}

Figure \ref{fig:hyprank_model} shows the overall architecture of the hypothesis reranker that is similar to \cite{YBKim2018b}.
First, we run intent classification and slot filling for the $k$ most confident domains from the shortlister outputs to obtain additional information for those domains \cite{Tur2011}.\footnote{Maximum Entropy model and Conditional Random Field model \cite{Lafferty2001} are utilized for intent classification and slot-filling, respectively.} Then, we compose $k$ hypotheses, each of which is a vector consists of the shortlister confidence score, intent score, Viterbi score of slot-filling, domain vector, intent vector, and the summation of the slot vectors. On top of the $k$ hypothesis vectors, a BiLSTM is utilized for representing contextualized hypotheses and a shared feed-forward neural network is used to obtain final confidence score for each hypothesis. We set $k$=3 in our experiments following \cite{YBKim2018b}.
We leverage the given ground-truth and the derived pseudo labels from the shortlister at the epoch showing the best dev set performance as target labels for training the reranker. We use hinge loss with margin 0.4 as the loss function.

One issue of the hypothesis reranking is that a training utterance cannot be used if no ground-truth exist in the top $k$ predictions of the shortlister.
This is problematic in the multi-label PU setting since correct domains can indeed exist in the top $k$ list but unknown, which makes the training utterance less useful in the reranking.
Our pseudo labeling method can address this issue. If correct pseudo labels are derived from the shortlister's top predictions for such utterances, we can use them properly in the reranker training, which was unavailable without them. This allows our approach make more improvement in hypothesis reranking than shortlisting.

\section{Experiments}
\label{sec:experiments}
In this section, we show training and evaluation sets, and experiment results.
\subsection{Datasets}
\label{ssec:datasets}
We utilize utterances with explicit invocation patterns from an intelligent conversational system for the model training similarly to \cite{YBKim2018b} and \cite{JKKim2018b}. For example, given ``ask \texttt{\{AmbientSounds\}} to \{\textit{play thunderstorm sound}\}'', we extract ``\textit{play thunderstorm}'' as the input utterance and \texttt{Ambient\\Sounds} as the ground-truth. One difference from the previous work is that we utilize utterances with positive system responses as the positive train set and the dev set, and use those with the negative responses as the negative train set as described in Section \ref{ssec:neg_feed_learn}.
We have extracted 3M positive train, 400K negative train, and 600K dev sets from 4M log data with 2,500 most frequent domains as the ground-truths. Pseudo labels are added to 53K out of 3M in the positive train set as described in Section \ref{ssec:pseudo_labeling}.

For the evaluation, we have extracted 10K random utterances from the user log data and independent annotators labeled the top three predictions of all the evaluated models for each utterance so that we can correctly compute nDCG at rank position 3.

\subsection{Experiment Results}
Table \ref{tab:experiments} shows the evaluation results of the shortlister and the hypothesis reranker with the proposed approaches. For the shortlisters, we show nDCG$_3$ scores, which are highly correlated with the F1 scores of the rerankers than other metrics since the second and third top shortlister predictions contribute the metric.
We find that just using the pseudo labels as the additional targets degrades the performance (2). However, when both the pseudo labels and the negative ground-truths are utilized, we observe significant improvements for both precision and recall (5). In addition, recall is increased when self-distillation is used, which achieves the best F1 score (6).
Each of utilizing the negative feedback $((1)\rightarrow(3) \;\text{and}\; (2)\rightarrow(5))$ and then additional pseudo labels $((3)\rightarrow(5) \;\text{and}\; (4)\rightarrow(6))$ show statistically significant improvements with McNemar test for p=0.05 for the final reranker results.

Using self-distillation $((3)\rightarrow(4) \;\text{and}\; (5)\rightarrow(6))$ shows increased F-1 score by increasing recall and decreasing precision, but the improvements are not significant. One issue is that pseudo labeling and self-distillation are contrary since the former encourages entropy minimization \cite{Grandvalet2005, Lee2013} while the latter can increase entropy by soft targeting the non-target labels. More investigation of self-distillation along with the proposed pseudo labeling would be future work.

Table \ref{tab:examples} shows examples of derived pseudo labels from model (6). It demonstrates that the domains capable of processing the utterances can be derived, which helps more correct model training.

\section{Conclusion}
We have proposed deriving pseudo labels along with leveraging utterances with negative system responses and self-distillation to improve the performance of domain classification when multiple domains are ground-truths even if only one ground-truth is known in large-scale domain classification.
Evaluating on the test utterances with multiple ground-truths from an intelligent conversational system, we have showed that the proposed approach significantly improves the performance of domain classification with hypothesis reranking.

As future work, combining our approach with pure semi-supervised learning, and the relation between pseudo labeling and distillation should be further studied.

\bibliographystyle{IEEEbib}
{\normalsize \bibliography{icassp2020}}

\end{document}